\documentclass[10pt,twocolumn,letterpaper]{article}

\usepackage{btas}
\usepackage{times}
\usepackage{epsfig}
\usepackage{graphicx}
\usepackage{amsmath}
\usepackage{amssymb}

\usepackage{subcaption}
\renewcommand{\vec}[1]{\mathbf{#1}}
\usepackage{color}

\btasfinalcopy 

\ifbtasfinal\fi

\usepackage{fancyhdr}
\fancypagestyle{empty}{%
  \fancyhf{}
  \chead{Accepted by IJCB 2017}
}
\begin{document}

\title{Deep Person Re-Identification with Improved Embedding and Efficient Training}

\author{Haibo Jin, Xiaobo Wang, Shengcai Liao\thanks{Corresponding author.}, and Stan Z. Li\\
Center for Biometrics and Security Research \& National Laboratory of Pattern Recognition\\
Institute of Automation, Chinese Academy of Sciences, Beijing, China\\
{\tt\small haibo.nick.jin@gmail.com}, {\tt\small \{xiaobo.wang, scliao, szli\}@nlpr.ia.ac.cn}}

\maketitle
\thispagestyle{empty}

\begin{abstract}
Person re-identification task has been greatly boosted by deep convolutional neural networks (CNNs) in recent years. The core of which is to enlarge the inter-class distinction as well as reduce the intra-class variance. However, to achieve this, existing deep models prefer to adopt image pairs or triplets to form verification loss, which is inefficient and unstable since the number of training pairs or triplets grows rapidly as the number of training data grows. Moreover, their performance is limited since they ignore the fact that different dimension of embedding may play different importance. In this paper, we propose to employ identification loss with center loss to train a deep model for person re-identification. The training process is efficient since it does not require image pairs or triplets for training while the inter-class distinction and intra-class variance are well handled. To boost the performance, a new feature reweighting (FRW) layer is designed to explicitly emphasize the importance of each embedding dimension, thus leading to an improved embedding. Experiments \footnote{See the code on https://github.com/jhb86253817/tf-re-id} on several benchmark datasets have shown the superiority of our method over the state-of-the-art alternatives on both accuracy and speed.
\end{abstract}

\section{Introduction}

Person re-identification aims to re-identify a query person across multiple non-overlapping cameras. The task is challenging since pedestrian images from different camera views suffer from large variations in  poses, lightings and backgrounds. Many earlier works solve the re-identification problem by dividing it into two separated parts: feature extraction~\cite{Liao15a, Liao15b, Zhao14, Yang14, Khamis14, Li13, Ma12, Xiong14} and metric learning~\cite{Paisitkriangkrai15, Zhang16a, Zhang16b, Chen16, Kodirov16, Liao15a, Liao15b}. A large number of hand-crafted features are designed to enhance the robustness of pedestrian images to pose, viewpoint and illumination changes. After the feature is extracted, metric learning is applied to learn a metric for the features so that the images of the same person are close while the ones of different pedestrians are far away from each other in the metric space.

\begin{figure}
\begin{center}
\includegraphics[width=0.9\linewidth]{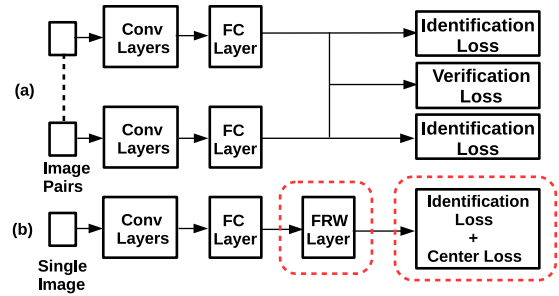}
\end{center}
   \caption{The difference between the state-of-the-art CNN \cite{Geng16,Zheng16b} and our proposed CNN for person re-identification. (a) The current best CNN has two branches, which takes a pair of images as input. (b) Our proposed CNN does not require image pairs or triplets for training since it utilizes the combination of identification loss and center loss. Moreover, a new feature reweighting (FRW) layer is designed so that the importance of each embedding dimension can be adaptively adjusted.}
   \label{CNN} \vspace{-5pt}
\end{figure}

In recent years, convolutional neural networks (CNNs) have achieved promising results on person re-identification~\cite{Ahmed15, Cheng16, Ding15, Geng16, Li14, Liu16, Shi16, Shi15, Ustinova15, Varior16a, Varior16b, Wang16, Xiao16a, Yi14, Zheng16b, Barbosa17, Zheng17} due to their advantages on feature learning. Different from previous works, CNNs learn features and metrics jointly from data in an end-to-end manner. Then an embedding is learned to measure the similarities between images using Euclidean distance. The loss function of a CNN plays an important role on its performance.  Verification loss~\cite{Ahmed15, Cheng16, Ding15, Li14, Shi16, Ustinova15, Varior16a, Varior16b, Wang16, Yi14} is popular among CNNs on person re-identification task benefited from its simple motivation: reducing the variance between intra-class embeddings while increasing the distinction between inter-class ones. However, verification loss takes image pairs or triplets for training, the number of which grows rapidly as the number of classes grows. When there are numerous person identities, verification loss is likely to show slow convergence and unstable performance. Identification loss is usually used for classification task, and it has also been applied to person re-identification task~\cite{Xiao16a, Zheng16a, Xiao16b} due to its simplicity and discriminative ability. Though identification loss can separate inter-class embeddings efficiently, it does not explicitly reduce the intra-class variance. Thus, the performance may be limited since the embeddings of the same person can have large distance on test data due to viewpoint, pose and background variations. To absorb the merits of the above two losses, recent works \cite{Zheng16b, Geng16} tend to combine them (please see Fig. \ref{CNN}a), and have achieved promising results. However, the inefficiency issue from verification loss still remains despite their performance improvement.

\begin{figure}
    \centering
    \begin{subfigure}[b]{0.22\textwidth}
        \includegraphics[width=\textwidth]{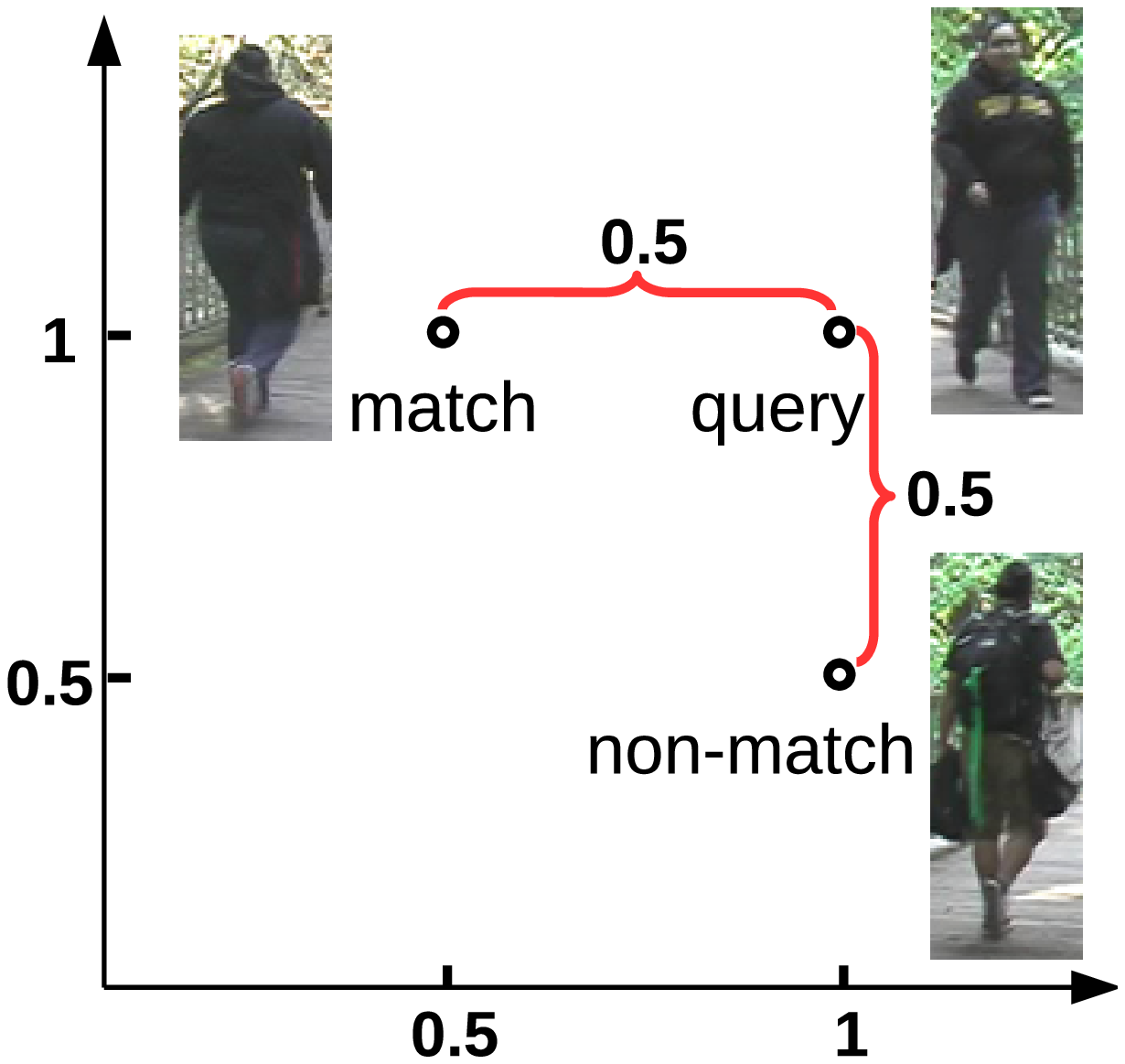}
        \caption{without FRW layer}\label{NFR}
    \end{subfigure}
    \begin{subfigure}[b]{0.22\textwidth}
        \includegraphics[width=\textwidth]{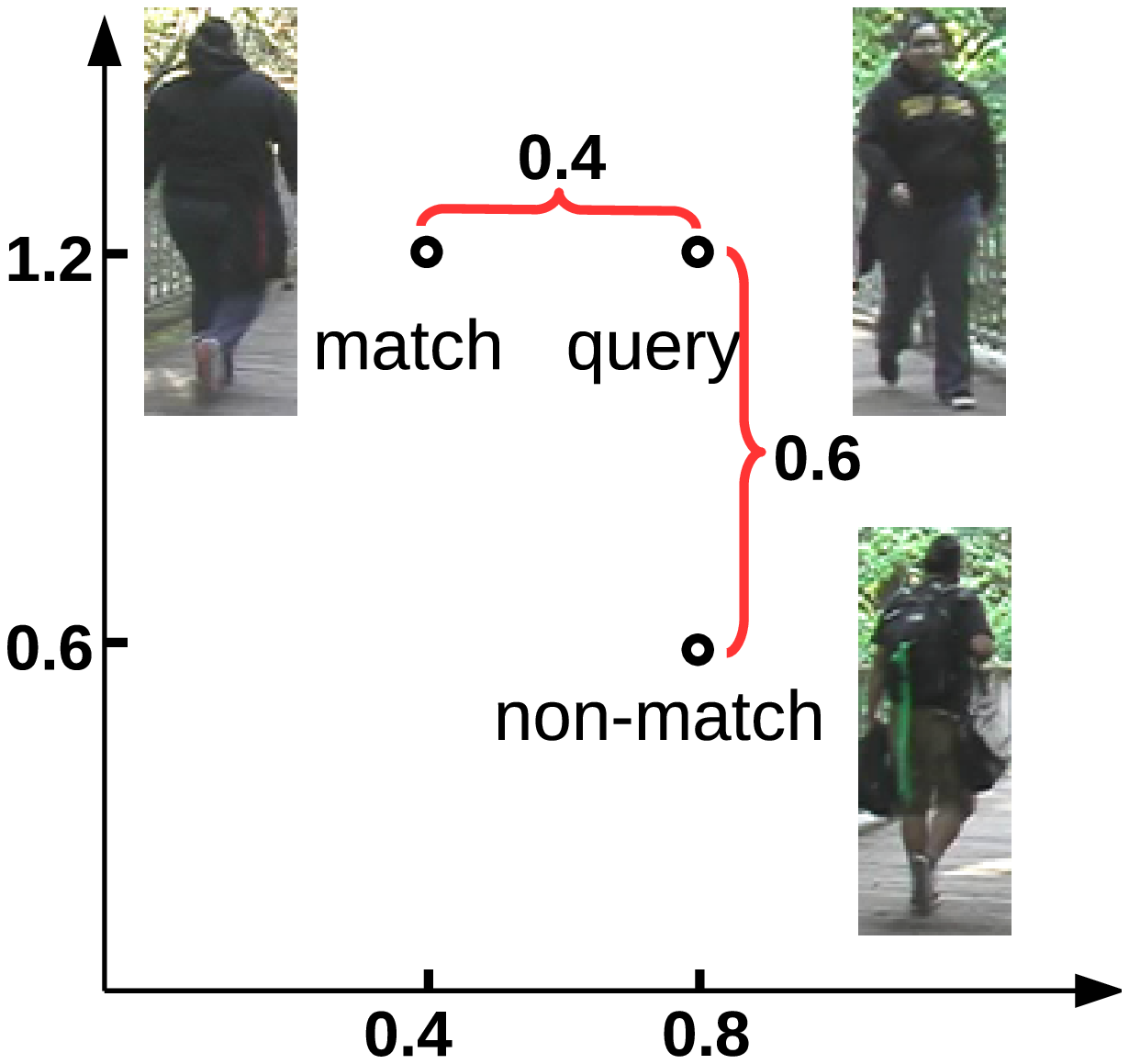}
        \caption{with FRW layer}\label{FR}
    \end{subfigure}
\caption{The effect of applying FRW layer. (a) Without FRW layer, the 2D embeddings of a matched image and a non-matched image have the same distance to the query embedding. (b) With FRW layer, the distance between the matched embedding and the query is closer than the non-matched one because the more essential dimension is enlarged while the less important one is shrinked.}
\end{figure}

In addition to the inefficiency problem, the existing deep embedding models draw little attention to the importance of each embedding dimension. They simply accumulate the squared difference of each dimension (\ie Euclidean distance) to measure the distance between embeddings~\cite{Cheng16, Ding15, Geng16, Varior16a, Varior16b, Xiao16a, Zheng16b}. In other words, each dimension of an embedding contributes equally to the total distance. Imagine that a matched embedding and a non-matched embedding have the same distance to the embedding of the query image (Fig. \ref{NFR}), Euclidean distance method is not able to distinguish the matched one from the two. If a model learns to measure the importance of each dimension, then reweights the embeddings so that the important dimension is emphasized while the unimportant one is depressed (Fig. \ref{FR}), such problem can be alleviated. Unfortunately, few works have considered the importance of different embedding dimensions.

To overcome the above two shortcomings, this paper proposes a new CNN model for person re-identification. Specifically, we employ identification loss with center loss to train CNN, which does not require image pairs or triplets as input. Center loss~\cite{Wen16} aims to pull images to the corresponding class center so that the intra-class variance is reduced. It functions similarly as verification loss but the learning process is more efficient. Meanwhile, a new feature reweighting (FRW) layer to adaptively learn the importance of each dimension has been designed. The FRW layer is placed after the embedding layer, performing element-wise multiplication upon its input. By doing so, the model gains the freedom to explicitly adjust the scales of the learned embeddings so that some less important features could be squeezed to avoid overfitting. Fig. \ref{CNN}b shows the structure of our proposed CNN. The contributions of this paper can be summarized as follows:
\begin{itemize}
\item {We employ identification loss with center loss to train a deep CNN model without constructing image pairs or triplets as input, thus improving the training efficiency.}

\item{We design a new FRW layer to explicitly emphasize the importance of each embedding dimension, leading to an improved embedding to boost the performance.}

\item{Experiments on CUHK03~\cite{Li14}, CUHK01~\cite{Li13} and VIPeR~\cite{Gray07} have validated the superiority of our method over the state-of-the-arts.}
\end{itemize}

\begin{figure*}
\begin{center}
\includegraphics[height=43mm,width=0.6\linewidth]{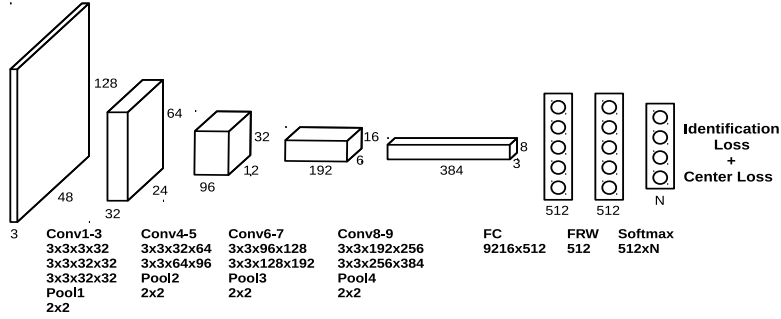}
\end{center}
   \caption{Our CNN architecture. A single image passes through several convolutional layers and max pooling layers, and a 512D vector is obtained by a fully connected layer. Then, FRW layer reweights the vector to get an improved embedding. Finally, the architecture is equipped with both identification loss and center loss to train the deep model.}\label{Model}
\end{figure*}

\section{Related Work}

To learn effective embeddings, existing works can be classified into two categories: 1) Improving the deep CNN structure to learn discriminative embeddings; 2) Designing better loss functions for deep CNN training.

\textbf{CNN structure:} To improve the CNN embeddings, Li \etal~\cite{Li14} propose to jointly handle various variations with filter pairing component. Yi \etal~\cite{Yi14} design a Siamese CNN to handle the divided images to finally compute a merged similarity score between images. Ahmed \etal~\cite{Ahmed15} propose a cross-input neighborhood difference layer to capture local relationship between two images as well as a patch summary layer to summarize the features learned from the previous layers. Wang \etal~\cite{Wang16} propose a joint framework of single-image representation and cross-image representation to get a merged result. Xiao \etal~\cite{Xiao16a} propose a CNN that learns features from multiple domains. Cheng \etal~\cite{Cheng16} design a multi-channel parts-based CNN to learn both the global and the local features. Varior \etal~\cite{Varior16a} propose a gating function for CNN to emphasize fine common local patterns. Varior \etal~\cite{Varior16b} use Long Short-Term Memory to emphasize contextual information for learning features. Shi \etal~\cite{Shi16} propose a moderate positive sample mining method to learn a variation insensitive feature. Sun \etal~\cite{Sun17} propose to add an Eigenlayer before the last fully connected (FC) layer to learn decorrelated weight vectors. Although existing CNN structures have achieved promising results, they still suffer from learning inefficiency problem because of verification loss.

\textbf{Loss function:} Binary identification loss, contrastive loss and triplet loss are three main types of verification loss. CNNs with binary identification loss have been used by ~\cite{Li14, Ahmed15}.  They output a binary prediction, indicating whether the two images belong to the same identity or not. Many other deep models learn an embedding for each image, and compute the similarities between embeddings based on the Euclidean distance. The works~\cite{Varior16a, Varior16b} use contrastive loss to train a CNN, which requires a pair of image samples for training. The methods~\cite{Wang14, Cheng16, Wang16, Ding15, Hermans17} employ triplet loss or its variations with CNN, which requires image triplets during training. For simplicity, the approaches~\cite{Xiao16a, Xiao16b, Zheng16a} apply identification loss to person re-identification task since it learns discriminative features efficiently. The combination of identification loss and verification loss has been found effective on face recognition~\cite{Sun14}, and it also gives excellent performance on person re-identification~\cite{Zheng16b, Geng16}. Recently, the work~\cite{Wen16} propose center loss to reduce the intra-class variance on face recognition task, without constructing image pairs or triplets during training. However, for person re-identification, the mainstream loss to handle the intra-class variance is still verification loss.

\section{Our CNN Model}

\subsection{The Overall Architecture}

Our proposed CNN model is a single CNN (different from the previous Siamese CNNs) that consists of nine convolutional layers, four max pooling layers, one FC layer, one FRW layer, and finally a softmax classifier. Fig. \ref{Model} gives the detailed illustration of the model. All the convolutional layers use 3 $\times$ 3 filters, with stride 1 and zero paddings. The max pooling layers all have 2 $\times$ 2 filters with stride 2. Batch normalization is applied after each convolutional layer or FC layer to speed up the training. Then leaky rectified linear unit (LReLU) is used after these layers as the non-linear activation function. After the FRW layer, we get a 512D embedding equipped with identification loss and center loss.

\subsection{Identification Loss and Center Loss}

Identification loss aims to enlarge the inter-class distinction and is usually used for multi-class classification task. It can be formulated as follows:
\begin{equation}\label{Identification}
L_I= -\frac{1}{M} \sum_{i=1}^M \text{log} \frac{e^{\mathbf{W}_{y_i}^T \vec{x}_i + b_{y_i}}}{\sum_{j=1}^N e^{\mathbf{W}_j^T \vec{x}_i + b_j}},
\end{equation}
where $M$ is the batch size, $\vec{x}_i \in \mathbb{R}^D$ is the $i$-th embedding of the batch, and $y_i$ is the class label of the current input. $\mathbf{W}_k$ is the $k$-th column of the FC weights $\mathbf{W} \in \mathbb{R}^{D \times N}$, $b_k$ is the $k$-th item of the bias term $\vec{b} \in \mathbb{R}^N$, and $N$ is the number of categories.

Center loss is proposed by Wen \etal~\cite{Wen16} to reduce the intra-class variance for face recognition. It maintains a center point for each class, and keeps pushing each image embedding to its corresponding center so that the variations between image embeddings and their centers are small. It can be formulated as follows:
\begin{equation}\label{Center}
L_C = \frac{1}{2M} \sum_{i=1}^M ||\vec{x}_i-\vec{c}_{y_i}||^2_2,
\end{equation}
where $\vec{c}_{y_i} \in \mathbb{R}^D$ is the corresponding center of the embedding $\vec{x}_i$. Specifically, unlike other CNN parameters, the updating of the class centers $\vec{c}_{y_i}$ are additionally performed instead of backpropagation:
\begin{equation}
\begin{aligned}
\Delta\vec{c}_j^t &= \frac{\sum_{i=1}^M \delta(y_i=j)\cdot (\vec{c}_j^t-\vec{x}_i)}{1+\sum_{i=1}^M \delta(y_i=j)}, \\
\vec{c}_j^{t+1} &= \vec{c}_j^t - \alpha \cdot \Delta \vec{c}_j^t,
\end{aligned}
\end{equation}
where $\alpha$ is the learning rate of the centers ranging from $0$ to $1$, $\delta(condition)=1$ if the $condition$ is satisfied, otherwise $\delta(condition)=0$.

During training, the two losses are optimized jointly using the formula as:
\begin{equation}
\begin{split}
L &= L_I + \lambda L_C \\
  &= -\frac{1}{M} \sum_{i=1}^M \text{log} \frac{e^{W_{y_i}^T \vec{x}_i + b_{y_i}}}{\sum_{j=1}^N e^{W_j^T \vec{x}_i + b_j}}
  + \frac{\lambda}{2M} \sum_{i=1}^M ||\vec{x}_i-\vec{c}_{y_i}||^2_2,
\end{split}
\end{equation}
where $\lambda$ is a scalar to balance the two loss functions. As we can see from Eq. (4), the loss function of our model only involves batches of single image samples, which leads to the improvement of training efficiency over the existing deep person re-identification models.

\subsection{FRW Layer}
The importance of each embedding dimension has always been assumed to be equal in the existing works, ignoring the difference between different dimension. Here we argue that CNN should have the freedom to learn such difference. In this work, a new FRW layer is proposed to reweight the learned embedding of a CNN. More specifically, FRW layer performs an element-wise product of an embedding and the FRW weights, formulated as follows:

\begin{equation}
\hat{\vec{x}} = \vec{x} \odot \vec{w}_{frw},
\end{equation}
where $\vec{x} \in \mathbb{R}^D$ is a learned embedding, $\vec{w}_{frw} \in \mathbb{R}^D$ is the weights of FRW layer, and $\odot$ denotes element-wise product. Intuitively, FRW layer enlarges certain dimensions of the embedding while reducing the other ones to strengthen the more essential features so that the similarities between embeddings can be reflected more accurately by Euclidean distance. For example, the central area of a pedestrian image can be more important than the border areas. For stability, additional constraint on the weights of FRW layer has been developed:

\begin{equation}
L_F = \beta \cdot (\frac{1}{2}||\vec{w}_{frw}||_2^2-C)^2,
\end{equation}
where $\beta$ controls the importance of the constraint, and $C$ is a constant to constrain the norm of the weight vector.

From another perspective, FRW layer can be seen as a separated part from softmax classifier. Among deep embedding models, the weights of softmax classifier are usually discarded after training because these weights are trained specifically on training classes, which are useless for different testing classes. Nevertheless, the trained weights of softmax classifier do contain general knowledge that is irrelevant to classes. We can treat the trained softmax weights as two parts: one that contains knowledge specific to each training class and the other that learns general knowledge applicable to all classes. Accordingly, the softmax classifier and the FRW layer in our model handle the two kinds of knowledge respectively. Formally, the standard softmax weights can be decomposed as follows:

\begin{equation}
\begin{split}
W_j^T \vec{x}+b_j &= (\hat{W}_j \odot \vec{w}_{frw})^T \vec{x} + b_j \\
                  &= \hat{W_j}^T(\vec{x} \odot \vec{w}_{frw}) + b_j,
\end{split}
\end{equation}
where $W_j$ is the $j$-th column of a standard softmax classifier weight, $\hat{W}_j$ is the $j$-th column of the softmax classifier weight from our model, and $\vec{w}_{frw}$ is the weight of our FRW layer. From Eq. (7), we can see that the FRW layer and the softmax classifier in our model is equivalent to the standard softmax classifier. By separating a FRW layer from the standard softmax classifier, the learned general knowledge about feature importance could be merged into the embeddings, and so it is applicable in testing phase.

\subsection{Training}

There are two types of training paradigms in this paper: 1) For relatively large dataset (\eg CUHK03~\cite{Li14}), we simply train the model on its training set using stochastic gradient descent with mini-batches; 2) As for small datasets (\eg CUHK01~\cite{Li13}, VIPeR~\cite{Gray07}), we adopt a similar deep transfer learning method from~\cite{Geng16}. We pretrain the model on large person re-identification datasets (CUHK03~\cite{Li14}+Market1501~\cite{Zheng15}), then fine-tune it on the corresponding training set of small data. Note that a two-stepped fine-tuning strategy from~\cite{Geng16} is used in this paper to conduct a more effective transfer learning. After pretraining, the weights of the softmax classifier cannot be reused in the fine-tuning stage because the two datasets have different classes. Therefore, the softmax classifier weights should be replaced by a randomly initialized one with $N_s$ nodes, where $N_s$ is the number of classes of the small dataset. Then, we do first-step fine-tuning by freezing the other parameters and only training the newly added weights until the classifier converges. After that, we fine tune all the parameters altogether as the second step. The reason for the two-stepped tuning is to avoid the newly added weights to backpropagate harmful gradients to the pretrained weights of the previous layers.

\subsection{Testing}

Testing is simple and efficient in the deep embedding senario. We feed all the testing images to the CNN model to get an embedding for each of them. Then we normalize each embedding to an unit vector. Finally, we compute the Euclidean distance between all the pairs from two camera views to measure the cumulative match curve (CMC).

\section{Experiments}

\subsection{Datasets}

\noindent \textbf{CUHK03:} CUHK03~\cite{Li14} consists of 13164 images from 1360 identities. It provides two settings, one annotated by human (labeled) and the other annotated by detectors (detected). We adopt the latter setting since it is closer to practical scenarios. Following the protocol in \cite{Li14}, we do 20 random splits, wherein 1160 identities are for training, 100 identities are for testing. The evaluation is in single shot.

\noindent \textbf{CUHK01:} CUHK01~\cite{Li13} contains 971 identities with two camera views, and each identity from each view has two images. Following the setting in~\cite{Geng16}, we randomly select one image for each identity in each view for both training and testing images. Then 485 identities are randomly selected for training, and the remaining 486 are for testing. The evaluation is based on 10 random splits, in single shot.

\noindent \textbf{VIPeR:} VIPeR~\cite{Gray07} contains 632 identities with two camera views. Each identity from each view has one image. Half of the identities are used for training, and the other half are for testing. The evaluation is also based on 10 random splits, in single shot.

\subsection{Data Preparation}

To reduce overfitting, we conduct data augmentation on each dataset. Each training image is augmented by 2D random translation as in ~\cite{Ahmed15, Li14}. We sample three images with 2D translation for each training image as well as a horizontal reflection. Each image is resized to $128 \times 48$. The mean of each training data is subtracted respectively.

\begin{table}[h]
\begin{center}
\begin{tabular}{|l|c|c|c|}
\hline
Method & Rank 1 & Rank 5 & Rank 10 \\
\hline\hline
XQDA	~\cite{Liao15a} & 46.3 & 78.9 & 88.6 \\
MLAPG~\cite{Liao15b} & 51.2 & - & - \\
DNS~\cite{Zhang16a} & 54.7 & 84.8 & 94.8 \\
LSSCDL~\cite{Zhang16b} & 51.2 & - & - \\
Siamese LSTM~\cite{Varior16b} & 57.3 & 80.1 & 88.3 \\
IDLA~\cite{Ahmed15} & 45.0 & 76.0 & 83.5 \\
Gated S-CNN~\cite{Varior16a} & 61.8 & 80.9 & 88.3 \\
EDM~\cite{Shi16} & 52.0 & - & - \\
Joint Learning~\cite{Wang16} & 52.2 & - & - \\
CAN~\cite{Liu16} & 63.1 & 82.9 & 88.2 \\
CNN Embedding~\cite{Zheng16b} & 66.1 & 90.1 & 95.5 \\
Deep Transfer~\cite{Geng16}\text{*} & \textbf{84.1} & - & - \\
\hline
CNN-I & 75.0 & 92.1 & 95.9 \\
CNN-IV & 80.2 & 94.9 & 97.3 \\
CNN-IC & 80.2 & 96.1 & 97.9 \\
CNN-FC-IC & 79.8 & 95.6 & 97.6 \\
CNN-FRW-IC & 82.1 & \textbf{96.2} & \textbf{98.2} \\
\hline
\end{tabular}
\end{center}
\caption{Accuracy on CUHK03 (detected). \text{*}Deep Transfer~\cite{Geng16} uses ImageNet for pretraining, while CNN-IV is our implementation of ~\cite{Geng16} without ImageNet pretraining.}
\end{table}

\subsection{Models for Comparison}

We compare our proposed model to a number of the existing methods, including state-of-the-art ones. In order to have a systematic comparison, we also implement several baseline models. We name the proposed model (Fig. 3) as CNN-FRW-IC. We implement a version without FRW layer to check the effectiveness of FRW layer, named CNN-IC. We also have one where the FRW layer is replaced by a FC layer, named CNN-FC-IC. We add the extra FC layer to check if simply increasing the depth of the network improves accuracy. There is also a version that only uses identification loss without FRW layer, named CNN-I.

To our knowledge, ~\cite{Geng16} gives the best accuracy among the existing deep embedding models, using identification loss and verification loss (binary identification loss) as loss function. So we implement a Siamese CNN with the two losses under our framework without FRW layer, named CNN-IV.

\subsection{Training Settings}
Our models are implemented using TensorFlow. We use the Adam optimizer to update parameters, where the exponential decay rate for the 1st and 2nd moment estimates are 0.9 and 0.999, respectively. The number of training iterations is 25k. The initial learning rate is 0.001, decayed by 0.1 after 22k iterations. The batch size is set to 100. The weight decay is 0.001. As for the center loss, the updating rate of the centers are $\alpha=0.5$, and its balance coefficient is $\lambda=0.01$. The balance coefficient of FRW layer is $\beta=0.001$, and the constant $C$ is set to 200.

\begin{table}[h]
\begin{center}
\begin{tabular}{|l|c|c|c|}
\hline
Method & Rank 1 & Rank 5 & Rank 10 \\
\hline\hline
SCSP~\cite{Chen16} & 53.5 & 82.6 & 91.5 \\
LSSCDL~\cite{Zhang16b} & 42.7 & \textbf{84.3} & \textbf{91.9} \\
TMA~\cite{Martinel16} & 43.8 & - & 83.9 \\
$\ell$1 GL~\cite{Kodirov16} & 41.5 & - & - \\
Siamese LSTM~\cite{Varior16b} & 42.4 & 68.7 & 79.4 \\
Metric Ensemble~\cite{Paisitkriangkrai15} & 45.9 & 77.5 & 88.9 \\
DNS~\cite{Zhang16a} & 51.2 & 82.1 & 90.5 \\
IDLA~\cite{Ahmed15} & 34.8 & 63.6 & 75.6 \\
DGD~\cite{Xiao16a} & 38.6 & - & - \\
MCP-CNN~\cite{Cheng16} & 47.8 & 74.7 & 84.8 \\
Gated S-CNN~\cite{Varior16a} & 37.8 & 66.9 & 77.4 \\
EDM~\cite{Shi16} & 40.9 & - & - \\
Joint Learning~\cite{Wang16} & 35.8 & - & - \\
Deep Transfer~\cite{Geng16}\text{*} & \textbf{56.3} & - & - \\
\hline
CNN-I & 39.1 & 61.3 & 70.5 \\
CNN-IV & 47.2 & 72.6 & 82.3 \\
CNN-IC & 49.3 & 77.3 & 87 \\
CNN-FC-IC & 46.6 & 74.4 & 84.3 \\
CNN-FRW-IC & 50.4 & 77.6 & 85.8 \\
\hline
\end{tabular}
\end{center}
\caption{Accuracy on VIPeR. \text{*}Deep Transfer~\cite{Geng16} uses ImageNet for pretraining, while CNN-IV is our implementation of ~\cite{Geng16} without ImageNet pretraining.}
\end{table}

\begin{table}[h]
\begin{center}
\begin{tabular}{|l|c|c|c|}
\hline
Method & Rank 1 & Rank 5 & Rank 10 \\
\hline\hline
$\ell$1 GL~\cite{Kodirov16} & 50.1 & - & -\\
DNS~\cite{Zhang16a} & 69.1 & 86.9 & 91.8 \\
IDLA~\cite{Ahmed15} & 47.5 & 71.6 & 80.3 \\
DGD~\cite{Xiao16a} & 66.6 & - & - \\
MCP-CNN~\cite{Cheng16} & 53.7 & 84.3 & 91 \\
Deep Transfer~\cite{Geng16}\text{*} & \textbf{77.0} & - & -\\
\hline
CNN-I & 63.4 & 84.4 & 90.5 \\
CNN-IV & 74.4 & \textbf{91.3} & \textbf{95.0} \\
CNN-IC & 70.1 & 90.5 & 94.8 \\
CNN-FC-IC & 66.1 & 88.2 & 93 \\
CNN-FRW-IC & 70.5 & 90.0 & 94.8 \\
\hline
\end{tabular}
\end{center}
\caption{Accuracy on CUHK01. \text{*}Deep Transfer~\cite{Geng16} uses ImageNet for pretraining, while CNN-IV is our implementation of ~\cite{Geng16} without ImageNet pretraining.}
\end{table}

\subsection{Results on CUHK03}

From Table 1, we can see that the model with only identification loss gets the worst accuracy among the baseline models. The accuracy of our implementation of identification loss and verification loss is slightly worse than the one reported in~\cite{Geng16} as they use extra ImageNet data for pretraining. Identification loss with center loss gets the same accuracy as identification loss with verification loss, which verifies the effectiveness of center loss. With the new designed FRW layer, the performance can be further improved. In contrast, the performance drops a little when a naive FC layer is added, which indicates that simply adding more layers does not bring any improvement. Among the models that do not use extra training data, our proposed model CNN-FRW-IC achieves the best rank 1, rank 5 and rank 10 accuracy on CUHK03 (detected).

\begin{table}[h]
\begin{center}
\begin{tabular}{|l|c|c|c|}
\hline
Method & Rank 1 & Rank 5 & Rank 10 \\
\hline\hline
CNN-IV & 77.0 & 93.1 & 96.6 \\
CNN-IC & 80.7 & 95.8 & 97.7 \\
\hline
\end{tabular}
\end{center}
\caption{Accuracy on CUHK03 (detected) with only 5k training iterations.}
\end{table}

\subsection{Results on VIPeR and CUHK01}

The results of VIPeR and CUHK01 are shown in Table 2 and 3, respectively. Similar to CUHK03, the results of our implementation of identification and verification loss are not as good as ~\cite{Geng16} due to the lack of ImageNet pretraining. CNN-IC reaches a higher performance than CNN-IV on VIPeR but a worse performance on CUHK01. The model with FRW layer has an improved accuracy on both the two datasets, and it outperforms most of the existing models. Similarly, adding a FC layer reduces the accuracy.

\subsection{Comparison on Convergence Speed}

Intuitively, center loss is more efficient than verification loss on training since it constructs batches of single image samples as input instead of person pairs or triplets. We conduct a comparative experiment on the two types of losses to see how their convergence speed differs in practice. We reduce the training iterations to 5k, where the learning rate is decayed by 0.1 at 4k iterations. The other settings remain the same as before. From Table 4, we see that the model of center loss has a better performance than the verification model. It is worth noting that the model accuracy of center loss with 5k training iterations is slightly better than the one with 25k training iterations, indicating that it has converged with 5k iterations. On the contrary, the accuracy of verification model drops when the number of training iterations is reduced. Therefore, center loss does converge faster than verification loss. More importantly, when a larger person re-identification dataset is used, the efficiency gap between the two losses will be more significant.

\section{Conclusion}

In this paper, we have proposed a novel CNN architecture for person re-identification. The proposed architecture utilizes identification loss and center loss to jointly balance the intra/inter class distances. By using center loss, our model becomes more efficient compared to the one using verification loss. Our model also contains a new FRW layer that learns to reweight the learned embedding for each dimension. Thus, the network gains more freedom to distribute the importance for each dimension. Based on the experimental results on CUHK03, CUHK01 and VIPeR, our proposed CNN outperforms the state-of-the-art competitors in most cases.

\section*{Acknowledgements}

This work was supported by the National Key Research and Development Plan (Grant No.2016YFC0801003), and the Chinese National Natural Science Foundation Projects \#61672521, \#61473291, \#61572501, \#61502491, \#61572536.

{\small
\bibliographystyle{ieee}
\bibliography{my_paper}

\begin{thebibliography}{10}\itemsep=-1pt

\bibitem{Ahmed15}
E.~Ahmed, M.~Jones, and T.~Marks.
\newblock An improved deep learning architecture for person re-identification.
\newblock In {\em CVPR}, 2015.

\bibitem{Barbosa17}
I.~Barbosa, M.~Cristani, B.~Caputo, A.~Rognhaugen, and T.~Theoharis.
\newblock Looking beyond appearances: Synthetic training data for deep cnns in
  re-identification.
\newblock {\em arXiv preprint arxiv:1701.03153}, 2017.

\bibitem{Chen16}
D.~Chen, Z.~Yuan, B.~Chen, and N.~Zheng.
\newblock Similarity learning with spatial constraints for person
  re-identification.
\newblock In {\em CVPR}, 2016.

\bibitem{Cheng16}
D.~Cheng, Y.~Gong, S.~Zhou, J.~Wang, and N.~Zheng.
\newblock Person re-identification by multi-channel parts-based cnn with
  improved triplet loss function.
\newblock In {\em CVPR}, 2016.

\bibitem{Ding15}
S.~Ding, L.~Lin, G.~Wang, and H.~Chao.
\newblock Deep feature learning with relative distance comparison for person
  re-identification.
\newblock {\em Pattern Recognition}, 48(10):2993--3003, 2015.

\bibitem{Geng16}
M.~Geng, Y.~Wang, T.~Xiang, and Y.~Tian.
\newblock Deep transfer learning for person re-identification.
\newblock {\em arXiv preprint arxiv:1611.05244}, 2016.

\bibitem{Gray07}
D.~Gray, S.~Brennan, and H.~Tao.
\newblock Evaluating appearance models for recognition, reacquisition, and
  tracking.
\newblock In {\em IEEE International Workshop on PETS}, 2007.

\bibitem{Hermans17}
A.~Hermans, L.~Beyer, and B.~Leibe.
\newblock In defense of the triplet loss for person re-identification.
\newblock {\em arXiv preprint arxiv:1703.07737}, 2017.

\bibitem{Khamis14}
S.~Khamis, C.-H. Kuo, V.~Singh, V.~Shet, and L.~Davis.
\newblock Joint learning for attribute-consistent person re-identification.
\newblock In {\em ECCV}, 2014.

\bibitem{Kodirov16}
E.~Kodirov, T.~Xiang, Z.~Fu, and S.~Gong.
\newblock Person re-identification by unsupervised $\ell$1 graph learning.
\newblock In {\em ECCV}, 2016.

\bibitem{Li13}
W.~Li and X.~Wang.
\newblock Locally aligned feature transforms across views.
\newblock In {\em CVPR}, 2013.

\bibitem{Li14}
W.~Li, R.~Zhao, T.~Xiao, and X.~Wang.
\newblock Deep{R}e{ID}: Deep filter pairing neural network for person
  re-identification.
\newblock In {\em CVPR}, 2014.

\bibitem{Liao15a}
S.~Liao, Y.~Hu, X.~Zhu, and S.~Li.
\newblock Person re-identification by local maximal occurrence representation
  and metric learning.
\newblock In {\em CVPR}, 2015.

\bibitem{Liao15b}
S.~Liao and S.~Li.
\newblock Efficient psd constrained asymmetric metric learning for person
  re-identification.
\newblock In {\em ICCV}, 2015.

\bibitem{Liu16}
H.~Liu, J.~Feng, M.~Qi, J.~Jiang, and S.~Yan.
\newblock End-to-end comparative attention networks for person
  re-identification.
\newblock In {\em IEEE Transactions on Image Processing}, 2016.

\bibitem{Ma12}
B.~Ma, Y.~Su, and F.~Jurie.
\newblock A novel image representation for person re-identification and face
  verification.
\newblock In {\em BMVC}, 2012.

\bibitem{Martinel16}
N.~Martinel, A.~Das, C.~Micheloni, and A.~Roy-Chowdhury.
\newblock Temporal model adaptation for person re-identification.
\newblock In {\em ECCV}, 2016.

\bibitem{Paisitkriangkrai15}
S.~Paisitkriangkrai, C.~Shen, and A.~Hengel.
\newblock Learning to rank in person re-identification with metric ensembles.
\newblock In {\em CVPR}, 2015.

\bibitem{Shi16}
H.~Shi, Y.~Yang, X.~Zhu, S.~Liao, Z.~Lei, W.~Zheng, and S.~Li.
\newblock Embedding deep metric for person re-identification: A study against
  large variations.
\newblock In {\em ECCV}, 2016.

\bibitem{Shi15}
H.~Shi, X.~Zhu, S.~Liao, Z.~Lei, Y.~Yang, and S.~Li.
\newblock Constrained deep metric learning for person re-identification.
\newblock {\em CoRR}, 2015.

\bibitem{Sun14}
Y.~Sun, Y.~Chen, X.~Wang, and X.~Tang.
\newblock Deep learning face representation by joint
  identification-verification.
\newblock In {\em NIPS}, 2014.

\bibitem{Sun17}
Y.~Sun, L.~Zheng, W.~Deng, and S.~Wang.
\newblock Svdnet for pedestrian retrieval.
\newblock {\em arXiv preprint arXiv:1703.05693}, 2017.

\bibitem{Ustinova15}
E.~Ustinova, Y.~Ganin, and V.~Lempitsky.
\newblock Multiregion bilinear convolutional neural networks for person
  re-identification.
\newblock {\em CoRR}, 2015.

\bibitem{Varior16a}
R.~Varior, M.~Haloi, and G.~Wang.
\newblock Gated siamese convolutional neural network architecture for human
  re-identification.
\newblock In {\em ECCV}, 2016.

\bibitem{Varior16b}
R.~Varior, B.~Shuai, J.~Lu, D.~Xu, and G.~Wang.
\newblock A siamese long short-term memory architecture for human
  re-identification.
\newblock In {\em ECCV}, 2016.

\bibitem{Wang16}
F.~Wang, W.~Zuo, L.~Lin, D.~Zhang, and L.~Zhang.
\newblock Joint learning of single-image and cross-image representations for
  person re-identification.
\newblock In {\em CVPR}, 2016.

\bibitem{Wang14}
J.~Wang, Y.~Song, T.~Leung, C.~Rosenberg, J.~Wang, J.~Philbin, B.~Chen, and
  Y.~Wu.
\newblock Learning fine-grained image similarity with deep ranking.
\newblock In {\em CVPR}, 2014.

\bibitem{Wen16}
Y.~Wen, K.~Zhang, Z.~Li, and Y.~Qiao.
\newblock A discriminative feature learning approach for deep face recognition.
\newblock In {\em ECCV}, 2016.

\bibitem{Xiao16a}
T.~Xiao, H.~Li, W.~Ouyang, and X.~Wang.
\newblock Learning deep feature representations with domain guided dropout for
  person re-identification.
\newblock In {\em CVPR}, 2016.

\bibitem{Xiao16b}
T.~Xiao, S.~Li, B.~Wang, L.~Lin, and X.~Wang.
\newblock End-to-end deep learning for person search.
\newblock {\em arXiv preprint arxiv:1604.01850}, 2016.

\bibitem{Xiong14}
F.~Xiong, M.~Gou, O.~Camps, and M.~Sznaier.
\newblock Person re-identification using kernel-based metric learning methods.
\newblock In {\em ECCV}, 2014.

\bibitem{Yang14}
Y.~Yang, J.~Yang, J.~Yan, S.~Liao, D.~Yi, and S.~Li.
\newblock Salient color names for person re-identification.
\newblock In {\em ECCV}, 2014.

\bibitem{Yi14}
D.~Yi, Z.~Lei, S.~Liao, and S.~Li.
\newblock Deep metric learning for person re-identification.
\newblock In {\em ICPR}, 2014.

\bibitem{Zhang16a}
L.~Zhang, T.~Xiang, and S.~Gong.
\newblock Learning a discriminative null space for person re-identification.
\newblock In {\em CVPR}, 2016.

\bibitem{Zhang16b}
Y.~Zhang, B.~Li, H.~Lu, A.~Irie, and X.~Ruan.
\newblock Sample-specific svm learning for person re-identification.
\newblock In {\em CVPR}, 2016.

\bibitem{Zhao14}
R.~Zhao, W.~Ouyang, and X.~Wang.
\newblock Learning mid-level filters for person re-identification.
\newblock In {\em CVPR}, 2014.

\bibitem{Zheng15}
L.~Zheng, L.~Shen, L.~Tian, S.~Wang, J.~Wang, and Q.~Tian.
\newblock Scalable person re-identification: A benchmark.
\newblock In {\em ICCV}, 2015.

\bibitem{Zheng16a}
L.~Zheng, H.~Zhang, S.~Sun, M.~Chandraker, Y.~Yang, and Q.~Tian.
\newblock Person re-identification in the wild.
\newblock {\em arXiv preprint arXiv:1604.02531}, 2016.

\bibitem{Zheng16b}
Z.~Zheng, L.~Zheng, and Y.~Yang.
\newblock A discriminatively learned cnn embedding for person
  re-identification.
\newblock {\em arXiv preprint arXiv:1611.05666}, 2016.

\bibitem{Zheng17}
Z.~Zheng, L.~Zheng, and Y.~Yang.
\newblock Unlabeled samples generated by gan improve the person
  re-identification baseline in vitro.
\newblock {\em arXiv preprint arXiv:1701.07717}, 2017.

\end{thebibliography}
}

\end{document}